\documentclass[11pt, a4paper, logo, twocolumn, copyright]{googledeepmind}

\usepackage[authoryear, sort&compress, round]{natbib}
\title{Perception Test 2026: Challenge Summary and Extension to City-scale Audio-Visual Reasoning}

\correspondingauthor{viorica@google.com}


\author[1]{Fedor Kitashov}
\author[1]{Jo\~ao Carreira}
\author[2]{Shiry Ginosar}
\author[1,3]{Dima Damen}
\author[1,4]{Andrew Zisserman}
\author[1]{Viorica P\u atr\u aucean}

\affil[1]{Google DeepMind}
\affil[2]{Toyota Technological Institute at Chicago}
\affil[3]{University of Bristol}
\affil[4]{University of Oxford}

\begin{abstract}
Continuing the Perception Test challenge series, we organised the fourth edition as a workshop at the European Conference on Computer Vision (ECCV) 2026 in Malmö, Sweden. This edition focused on spatial intelligence and featured four different tracks: unified multiple-choice videoQA and grounded videoQA from the original Perception Test benchmark, alongside two new tracks based on city-scale walking-tour videos (KilometerAudio and KilometerVision). In this report, we describe the new benchmarks used for city-scale tracks and summarise the winning solutions across all tracks, including a generalist model that competed across all tracks with satisfactory performance. The winning solutions in the newly added city-scale tracks demonstrated that complex spatial and multimodal reasoning can be solved by expensive agentic pipelines, but remains difficult for multimodal models used standalone.

\end{abstract}

\begin{document}

\maketitle

\section{Introduction}
Frontier AI models have demonstrated super-human capabilities in multiple areas, from  solving Millennium problems in Mathematics~\citep{openai2026navierstokesblog}, to controlling plasma for nuclear fusion reactors with superhuman precision~\citep{torax2024arxiv}, and writing complex software at speed much higher than humans. Yet, the general perception capabilities of these models are still falling short compared to average humans, especially on spatial understanding tasks. This might not come as a surprise, given that these models are mainly trained with passive internet-scale data, whereas the common dogma states that environment interaction is essential for learning perception~\citep{poincare2011science,held1963movement}.       

In this context, we dedicated the fourth edition of the Perception Test challenge\footnote{\url{https://perception-test-challenge.github.io/}} to probing spatial intelligence in frontier models across different environment scales. We rely on the original Perception Test benchmark to evaluate small-scale (table-top) spatial capabilities. For large-scale spatial understanding, we leverage the recently introduced KilometerVision benchmark~\citep{mahendran2026kilometer}, together with a newly created benchmark for audio-visual reasoning in large-scale environments; see details in sections~\ref{sec:kmvision} and ~\ref{sec:kmaudio}.

\noindent \textbf{Challenge tracks:} The challenge featured four tracks: Unified VideoQA, Grounded VideoQA, KilometerVision, and KilometerAudio. The first two tracks relied on videos from the original Perception Test benchmark, whereas the latter two tracks relied on publicly available Youtube Walking-Tour videos, annotated specifically for each benchmark.

\begin{figure*}[t!]
    \centering
    \includegraphics[width=0.9\textwidth]{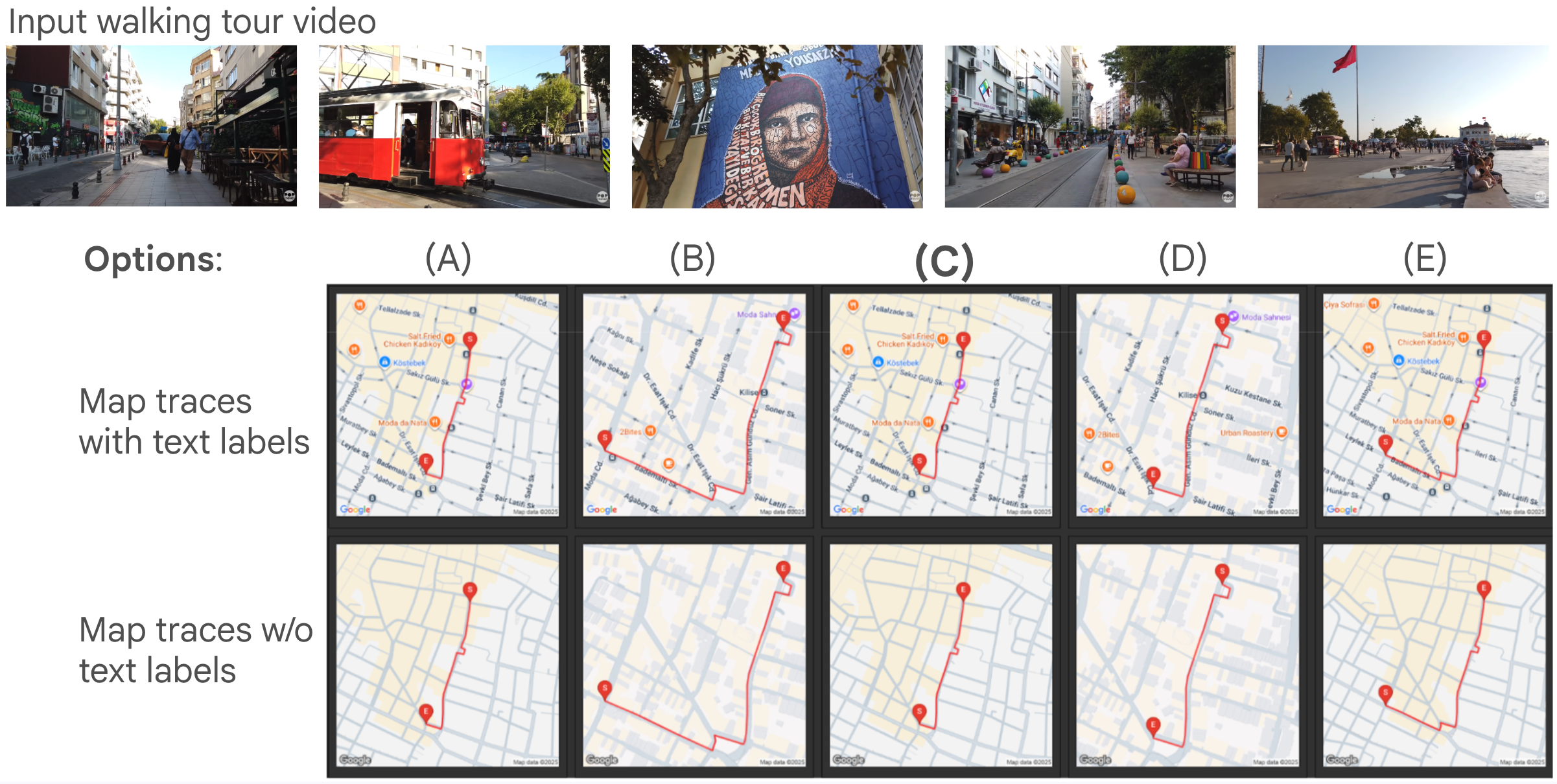}
    \caption{Example of map trace tasks in the KilometerVision benchmark. The model receives a video and a text question (\textit{``Which of these maps correctly represent the path of the person in this video?''}) together with options given as raster images, with or without text labels inpainted on the maps.}
    \label{fig:maps}
\end{figure*}

\begin{figure*}[h]
    \centering
    \includegraphics[width=0.9\textwidth]{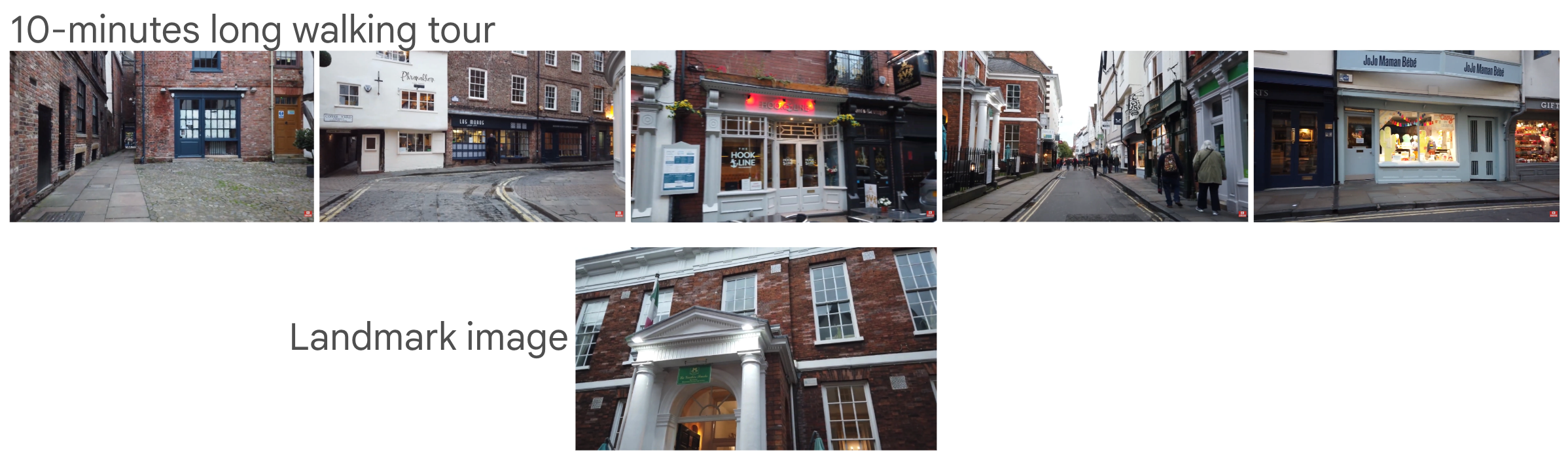}
    \caption{Example of landmark task in the KilometerVision benchmark, where the question contains a reference image in addition to the text prompt (\textit{``Does the image show a location in the video, and if so, what is the straight-line
distance between the camera location of the image and the camera location of the
final video frame?''}).}
    \label{fig:landmark}
\end{figure*}

\noindent \textbf{Challenge setup and prizes:}
The competition was hosted on the EvalAI platform. Each track had a validation and a test phase. The validation servers opened in mid-June 2026, and the test servers became available starting in mid-July (for unified and grounded VQA) and mid-August 2026 (for KilometerAudio and KilometerVision). The submission period closed at the end of August 2026. Participants were required to report the evaluation mode (zero-shot, few-shot, or fine-tuned) and submit a short technical report. Only the test submissions made public on the leaderboard were eligible for awards with monetary prizes, which totalled 22K EUR across tracks. We awarded runner-up and winners in each track, together with a best generalist award for an entry that competed across all tracks with the same backbone and obtained strong performance. We also offered honorary best performance certificates in KilometerVision and KilometerAudio tracks to a team that obtained the best scores in these tracks, but was not eligible for awards due to Google affiliation.

\section{KilometerVision Benchmark}
\label{sec:kmvision}
KilometerVision~\citep{mahendran2026kilometer} is a recent benchmark that probes city-scale spatial capabilities in frontier models using real-world walking tour Youtube videos, up to 10 minutes long, corresponding to about 1KM walking distance. The tasks use multiple-choice videoQA format, with three different task configurations: standard multiple-choice video QA with five text options; multiple-choice video QA where the five options are images (e.g.\ candidate route maps, see example in Figure~\ref{fig:maps}); and multiple-choice video QA where the question includes an image (e.g.\ a landmark whose distance to the final camera position must be estimated, see example in Figure~\ref{fig:landmark}). The benchmark includes seven types of questions: landmark recognition, distance to landmark, compass, loop closure, route summary, map trace (with or without text labels on the map), and Euclidean distance. Please check the original publication~\citep{mahendran2026kilometer} for more details about the tasks and data collection.

\begin{figure*}[t]
    \centering
    \includegraphics[width=0.98\textwidth]{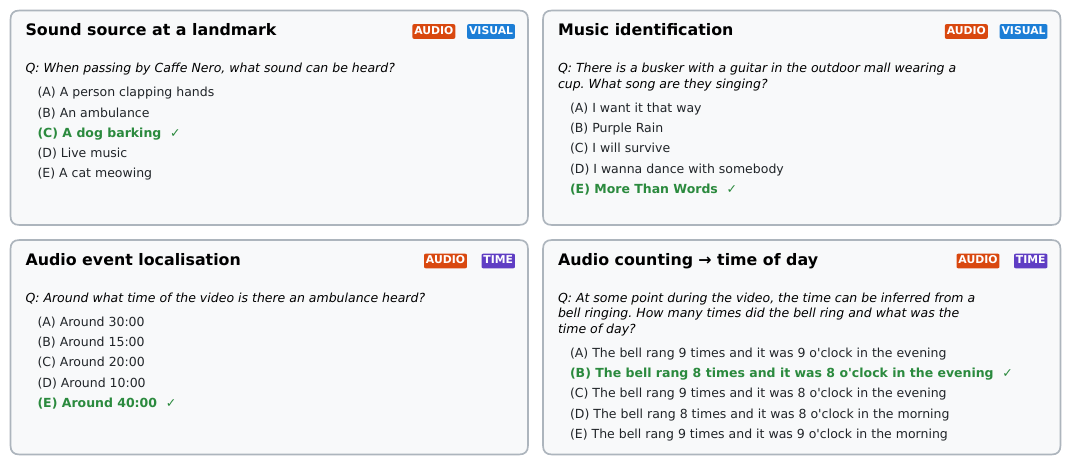}
    \caption{Example KilometerAudio questions from the validation set (correct answer in green). Each requires reasoning over the audio track of an hour-long walking-tour video.}
    \label{fig:kaexamples}
\end{figure*}

\begin{figure}[t]
    \centering
    \includegraphics[width=0.98\linewidth]{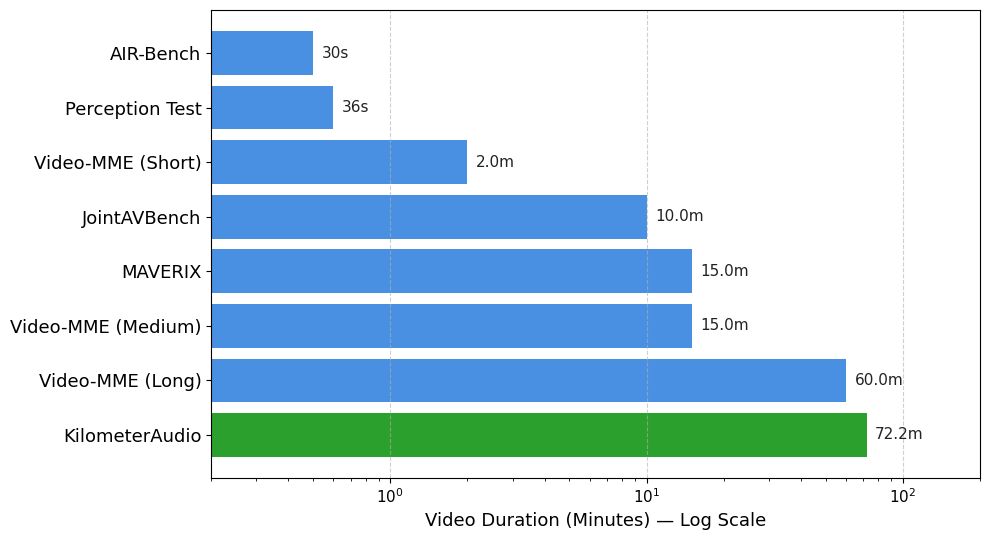}
    \caption{Multiple-choice audio-video benchmarks. The proposed KilometerAudio benchmark has the longest videos (more than 70 minutes on average) and the questions focus exclusively on audio events; Video-MME (Long) has videos up to 1h long, but only a fraction of the questions focus on audio events.}
    \label{fig:audiobench}
\end{figure}

\section{KilometerAudio Benchmark}
\label{sec:kmaudio}

To assess models' audio-visual capabilities in large-scale environments, we introduce the KilometerAudio benchmark. 
This is a novel benchmark, introduced here for the first time.
It aims to test the ability of foundational models, including agentic approaches, to reason about the interlink between audio and visual modalities in very long videos -- i.e. videos captured over walking for several kilometers around an urban environment. Natural sounds that are automatically captured in these public YouTube videos include sounds of human-made objects (cars, motorbikes, church bells, etc.), natural sounds (birds chirping and dogs barking), speech as well as acoustic urban pollution. Importantly, the videos are not edited to introduce irrelevant sounds (e.g. overlay music), making them ideal for multimodal reasoning. Due to the nature of walking videos, audio is often caused by an event `out of sight' of the camera.
Reasoning about an in-view visual along with an out-of-view audio as well as the length of the videos make this benchmark particularly unique and challenging.


Compared to existing audio-video benchmarks, e.g.~\citep{yang-etal-2024-air,patraucean2023perception,Fu_2025_CVPR,xie2025maverix,chao2026jointavbench}, KilometerAudio uses significantly longer videos (see Figure~\ref{fig:audiobench} for a comparison) and focuses exclusively on audio events.   
The test set contains 500 five-option questions over 218 publicly available hour-long Youtube walking-tour videos (1--3\,h). Video-MME Long has videos up to 1h long, but only a small fraction of the questions focus on audio events.

All the questions in KilometerAudio were collected with human raters and checked by our research team. The raters were instructed to identify recognisable sounds in the videos and formulate questions that require both the audio and visual modalities to be answered: identifying sound sources in space, localising or timing audio events, counting them, recognising songs or speech; see Figure~\ref{fig:kaexamples}. The raters were also required to provide the answers to the questions, together with plausible difficult negative options. The validation split (8 questions on 6 videos) reuses the audio-related questions of the hour-long video QA benchmark of the 2024 and 2025 editions~\citep{heyward2024perceptiontest2024challenge}. All test questions are new, and only 3 of the 218 test videos were used in previous editions. No training data are provided, so the track is zero-shot; see Table~\ref{tab:ka}.

\begin{table}[h]
    \centering
    \small{
    \begin{tabular}{lrr}
    \toprule
      \textbf{Split} & \textbf{Num videos} & \textbf{Num questions} \\
      \midrule
      Train & -- & -- \\
      Validation & 6 & 8 \\
      Test & 218 & 500 \\
      \bottomrule
    \end{tabular}}
    \caption{Splits of the KilometerAudio track.}
    \label{tab:ka}
\end{table}

\section{Overall challenge summary}

\begin{figure*}[t]
    \centering
    \begin{minipage}[t]{0.485\textwidth}
      \centering
      \includegraphics[height=5.6cm]{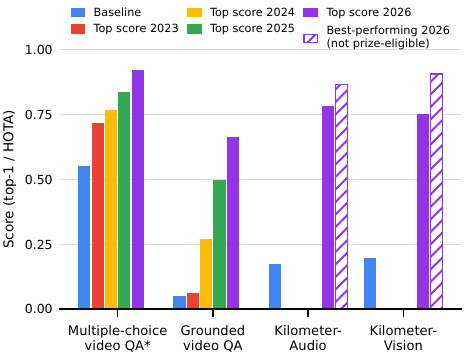}
      \caption{Per-track performance of the 2026 winners compared to baselines and to the best models of previous years, all on the same test sets. *For multiple-choice video QA we show accuracy on the 11,528 original questions, the only test questions common to all editions. KilometerAudio and KilometerVision were introduced in 2026; striped bars show the best-performing but non-prize eligible entry.}
      \label{fig:summary2026}
    \end{minipage}\hfill
    \begin{minipage}[t]{0.485\textwidth}
      \centering
      \includegraphics[height=5.6cm]{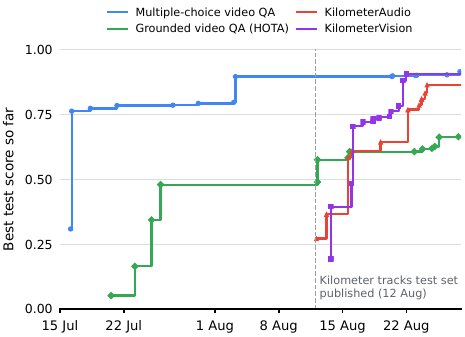}
      \caption{Best test score in each track during the 2026 test phase. The test servers for KilometerAudio and KilometerVision went live in August 2026, pending data approvals.}
      \label{fig:pertask}
    \end{minipage}
\end{figure*}



We received 1254 test submissions from 120 different teams across the four tracks. As it can be observed in Figure~\ref{fig:summary2026}, the performance improved steadily across editions on the unified and grounded videoQA tracks (using videos from the original Perception Test benchmark): from 0.812 (in 2025) to 0.914 top-1 accuracy in unified multiple-choice video QA in 2026, and from 0.497 (in 2025) to 0.664 HOTA in grounded video QA in 2026. In the two new tracks, the best-performing entry (non-eligible for prizes) reached close to perfect performance (0.864 on KilometerAudio and 0.905 on KilometerVision). The best prize-eligible submissions were also very strong, reaching 0.782 top-1 accuracy on KilometerAudio and 0.753 for KilometerVision. 

Figure~\ref{fig:pertask} shows the best test score of each track over the test phase. Due to data availability, the test servers for the two new tracks went public only later during August 2026.


\section{Challenge Tracks, Results, Awards}
For each track, we summarise the 2026-specific setup, the results, and the awarded methods. Team reports, with author names and affiliations, are linked from the workshop website: \url{https://perception-test-challenge.github.io/}.

\subsection{Unified multiple-choice video QA}

\noindent \textbf{Task, data, and metric.} These are unchanged from 2025; see \citet[Sec.~2 and 4.1]{heyward2026perceptiontest2025challenge}. Specifically, the test set combines the 11,528 original 3-option questions with the 1,842 5-option unified questions (point tracking, action ordering, object--action interaction, pretend action), which use visual prompts to recast tracking and temporal localisation as questions; the unified questions have no training split  enforcing zero-shot evaluation. The metric is top-1 accuracy.

\noindent \textbf{Baselines:} A random baseline scores 0.313 on the test set (chance level 31.5\%, i.e.\ 33.3\% on the 3-option and 20\% on the 5-option questions, weighted by their counts).


\begin{figure}[t]
    \centering
    \includegraphics[width=\linewidth]{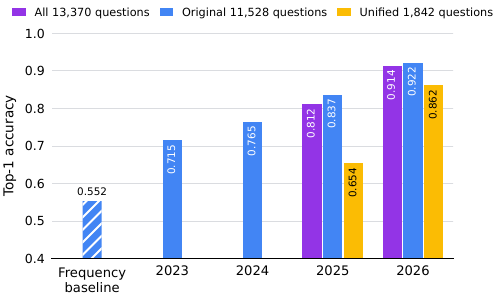}
    \caption{Best multiple-choice video QA test results per edition. The 11,528 original questions are common to all editions, and the 1,842 unified questions were added in 2025; 2023, 2024, and the frequency baseline (striped) are therefore on the original questions only, while for 2025 and 2026 we show the accuracy on all 13,370 questions and on each subset.}
    \label{fig:mcqa}
\end{figure}

\noindent \textbf{Results:}

\begin{table}[h]
    \centering
    \small{
    \begin{tabular}{llc}
    \toprule
       \textbf{Rank} & \textbf{Team name} & \textbf{top-1} \\
       \midrule
       Baseline & Random & 0.313 \\
       Runner-up & F423 (zero-shot) & 0.910 \\
       Best & Njust-KMG (fine-tuned) & 0.914 \\
         \bottomrule
    \end{tabular}}
    \caption{Multiple-choice video QA results on the test set.}
    \label{tab:mcqaresults}
\end{table}

\begin{figure*}[t]
    \centering
    \begin{minipage}[c]{0.60\linewidth}
      \includegraphics[width=\linewidth]{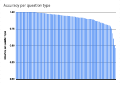}
    \end{minipage}\hfill
    \begin{minipage}[c]{0.36\linewidth}
      \includegraphics[width=\linewidth]{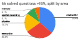}\\[4pt]
      \includegraphics[width=\linewidth]{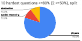}
    \end{minipage}
    \caption{Left: accuracy of the best 2026 submission (Njust-KMG) on each of the 141 unique questions. Right: distribution across areas of the questions that are quasi-solved (accuracy $>95\%$, top) and of the 10 hardest questions (accuracy $\leq 80\%$, bottom).}
    \label{fig:statsvqa}
\end{figure*}

Table~\ref{tab:mcqaresults} lists the awarded entries compared to the random baseline. The winning team, Njust-KMG, used a question-type-aware weighted ensemble of three models: Qwen3-VL-32B fine-tuned on the Perception Test training set (with a curriculum stage on the hardest question types and option shuffling to reduce position bias), and the Doubao-Seed-2.0-Pro and Gemini~3.1~Pro APIs queried zero-shot with question-type-specific prompts; the ensemble weights are derived from per-type validation accuracy. While the individual models obtain 78--83\% on the test set, the ensemble reaches 91.4\%. The runner-up, F423, used a single zero-shot model, Doubao-Seed-2.0-Lite, with evidence-first prompting (extracting the visual, temporal, and audio evidence before comparing the options), adaptive frame sampling (1--2 fps depending on the question type), and strict answer validation with at most one re-query, reaching 91.0\%. Please check the workshop website for more details on the methods included in the submission reports.

Relative to earlier editions (Figure~\ref{fig:mcqa}), the 2026 winner improved significantly on the questions added in 2025 (86.2\% vs 65.4\% for the best 2025 model) compared to improvement on the original set (92.2\% vs 83.7\%).  This indicates that the reformulated tracking and localisation questions are quickly becoming tractable for current models.

Figure~\ref{fig:statsvqa} breaks the winner's accuracy down by question type and shows which areas the quasi-solved ($>95\%$) and the hardest questions come from. Of the 141 questions, 66 (46.8\%) are now quasi-solved, compared with 36 (25.5\%) for the 2025 winner on the same questions, and only 2 questions obtain an accuracy of 50\% or lower (14 in 2025), both in the Physics area. Physics also dominates the 10 hardest questions (accuracy $\leq 80\%$; 8 Physics, 1 Spatial reasoning, 1 Abstraction). The remaining errors therefore concentrate on intuitive physics (e.g.\ judging the temperature of poured water, or predicting outcomes when imagining a change in velocity or friction of an object) rather than on recall or counting.

\subsection{Grounded video QA}

\noindent \textbf{Task, data, and metric.} These are identical to previous editions~\citep{heyward2026perceptiontest2025challenge}: given a video and a question, the model must track the object(s) that answer it, on the same 932 test videos (1,859 questions), evaluated with HOTA~\citep{luiten2020IJCV}. The baseline runs the MDETR detector~\citep{kamath2021mdetr} on the middle frame with the question as query and keeps the detections static throughout the video.

\noindent \textbf{Results:}
Table~\ref{tab:gqa} lists the two awarded entries compared to the baseline. Both solutions use a strong VLM to decide which objects answer the question and a SAM-family model for dense tracking. The winning team, Banana Bats, proposed Gemini-SAM3: Gemini~3.7~Flash, used zero-shot, builds a tracking plan (target identities, visibility intervals, anchor frames and boxes, checked by a critic) and provides sparse boxes at 1 fps; SAM~3~\citep{carion2025sam3}, prompted with the anchor boxes, propagates the targets bidirectionally, and its tracks are fused with the sparse Gemini tracks only when they pass reliability checks. A variant of SAM~3 with a small residual head trained for amodal (occlusion-aware) boxes on external data is selected for multi-target questions. The runner-up, Huawei SpaceMind, used GPT-5.6 Sol zero-shot to identify the answer objects and produce complete box tracks, then refined only the box coordinates, never the object identities, using a second independent track, repeated queries on close-up crops, and SAM~2.1~\citep{sam2} box refinement.

\begin{figure}
    \centering
    \includegraphics[width=\linewidth]{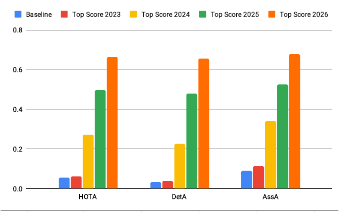}
    \caption{Baseline vs best results over time in terms of overall HOTA, detection, and association accuracy for the grounded video QA task.}
    \label{fig:hota}
\end{figure}

Figure~\ref{fig:hota} compares the top 2026 model to the best entries of previous editions: HOTA increased from 0.497 to 0.664, about one third higher, with similar gains in detection (0.479 to 0.654) and association accuracy (0.524 to 0.677). In 2023 the best entry barely exceeded the static baseline (0.063 vs 0.052 HOTA). Only three years later, zero-shot pipelines that pair a VLM with a promptable tracker produce very strong performance, showing the high pace of progress in the field.

\begin{table}[h]
    \centering
    \footnotesize{
    \setlength{\tabcolsep}{3pt}
    \begin{tabular}{llccc}
    \toprule
       \textbf{Rank} & \textbf{Team name} & \textbf{HOTA} & \textbf{DetA} & \textbf{AssA} \\
       \midrule
       Baseline & MDETR+static & 0.052 & 0.033 & 0.091 \\
       Runner-up & Huawei SpaceMind & 0.654 & 0.651 & 0.661 \\
       Best & Banana Bats & 0.664 & 0.654 & 0.677 \\
         \bottomrule
    \end{tabular}}
    \caption{Grounded video QA results on the test set.}
    \label{tab:gqa}
\end{table}

\subsection{KilometerVision}

\noindent \textbf{Task, data, and metric:} The model needs to answer 5-way multiple-choice questions that can have different configurations as mentioned in section~\ref{sec:kmvision}. The challenge test set contains 986 questions on 54 walking-tour video segments, spanning six question types: landmark recognition and distance to landmark (147), compass (146), loop closure (112), route summary (147), start--end Euclidean distance (150), and map trace with or without text labels on the map (284 questions); see Table~\ref{tab:kv}. No training data are provided. The evaluation metric is top-1 accuracy over the five options.

\begin{table}[h]
    \centering
    \small{
    \begin{tabular}{lrr}
    \toprule
      \textbf{Split} & \textbf{Num videos} & \textbf{Num questions} \\
      \midrule
      Train & -- & -- \\
      Validation & 13 & 17 \\
      Test & 54 & 986 \\
      \bottomrule
    \end{tabular}}
    \caption{Splits of the KilometerVision track.}
    \label{tab:kv}
\end{table}


\noindent \textbf{Baselines:} A random baseline obtains 0.196 on the test set (expected chance level 0.20).

\begin{figure}
    \centering
    \includegraphics[width=\linewidth]{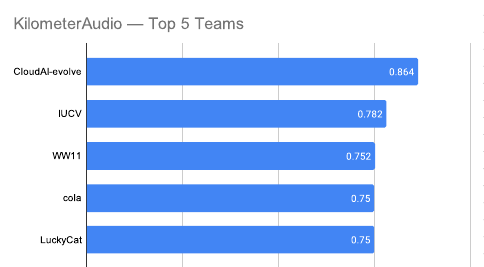}\\[4pt]
    \includegraphics[width=\linewidth]{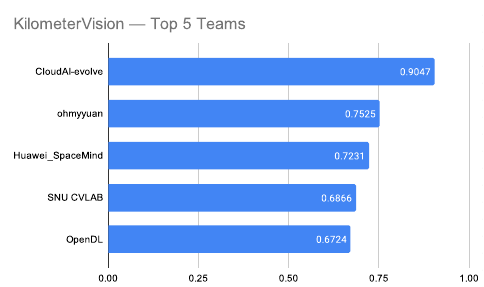}
    \caption{Top-5 public test results (top-1 accuracy) in KilometerAudio (top) and KilometerVision (bottom), including the best-performing entry, CloudAI-evolve.}
    \label{fig:kmtop5}
\end{figure}

\noindent \textbf{Results:} The results are included in Table~\ref{tab:kvwin} and Figure~\ref{fig:kmtop5}. The top solutions combine a VLM for perception and world-knowledge grounding with explicit geometric estimation. 
The winning team, ohmyyuan, proposed evidence-log video agents: Gemini~3.7~Flash builds a timestamped log of street names, heading changes, and landmarks from frame montages and keyframes, and answers with a family-specific playbook; self-consistency over several sessions is combined with deterministic estimators (optical-flow yaw odometry and visual place recognition) that can override the majority vote, and an independent verifier re-solves questions without a clear majority. The runner-up, Huawei SpaceMind, combined GPT-5.6 Sol for semantic reasoning with a metric camera path reconstructed with VGGT~\citep{wang2025vggt} and a metric-depth model; question-specific solvers (summed yaw for compass questions, start--end displacement for distances, and route-shape matching for map questions) decided 642 of the 986 answers. The CloudAI-evolve team, which was not eligible for a prize, obtained the best overall score (0.905) and was highlighted as the best-performing entry. Their approach tackled the task with an agentic pipeline with unlimited token budget that
explored the validation set and a small fraction
(about 10\%) of the unlabelled test questions to learn a strategy for solving the task on the full test set. The strategy involved calling metric 3D reconstruction models (VGGT-omega), object-size-based scale estimation tools, and grounding the videos on the map.  

\begin{table}[h]
    \centering
    \small{
    \begin{tabular}{llc}
    \toprule
       \textbf{Rank} & \textbf{Team name} & \textbf{top-1} \\
       \midrule
       Baseline & Random & 0.196 \\
       Runner-up & Huawei SpaceMind & 0.723 \\
       Best & ohmyyuan & 0.753 \\
       \midrule
       Best-performing$^\dagger$ & CloudAI-evolve & 0.905 \\
         \bottomrule
    \end{tabular}}
    \caption{KilometerVision results. $^\dagger$Best-performing entry, not eligible for a prize; see text.}
    \label{tab:kvwin}
\end{table}

\subsection{KilometerAudio}

\noindent \textbf{Task description:} In the KilometerAudio task, the model receives a full hour-long walking-tour video, including its audio track, together with a question and five possible answers, out of which only one is correct. The questions  require processing both the visual and the audio track of the video, e.g.\ identifying sounds, understanding speech, or correlating audio events with visual content, possibly through multi-hop reasoning.

\noindent \textbf{Metric:} The evaluation metric is top-1 accuracy, i.e.\ the percentage of questions where the predicted option (1 out of 5) matches the ground truth.

\noindent \textbf{Baselines:} A random baseline obtains 0.174 on the test set (expected chance level 0.20).

\noindent \textbf{Results:} The results are included in Table~\ref{tab:kawin} and Figure~\ref{fig:kmtop5}. 
The winning team, IUCV, proposed AVRA, a training-free audio-visual reasoning agent built on Gemini~3.7~Flash. For each video, a question-independent index is pre-computed with frozen models: audio-event scores for 447 AudioSet classes from BEATs~\citep{beats}, object detections, and multilingual OCR. At test time, a fresh agent session localises a few candidate time windows from the index, inspects the raw audio and video directly, keeps an explicit evidence log, and derives an answer before reading the options. The runner-up, WW11, proposed a zero-shot tool-using agent in which Gemini~3.1~Pro plans the search and reasons over at most 48 frames or crops per question, using offline indexes of timestamped speech transcripts (Whisper) and CLAP-based sound-event retrieval, and an audio model (Qwen3-Omni) to verify short audio excerpts; an answer is changed only when at least two independent modalities agree. The CloudAI-evolve team obtained again the best overall score (0.864) with a similar agentic pipeline as in the KilometerVision track, but using signal-processing tools instead of 3D reconstruction, to compile a strategy for solving the tasks; Gemini models were used for all reasoning steps. More details can be found on the challenge website.

\begin{table}[h]
    \centering
    \small{
    \begin{tabular}{llc}
    \toprule
       \textbf{Rank} & \textbf{Team name} & \textbf{top-1} \\
       \midrule
       Baseline & Random & 0.174 \\
       Runner-up & WW11 & 0.752 \\
       Best & IUCV & 0.782 \\
       \midrule
       Best-performing$^\dagger$ & CloudAI-evolve & 0.864 \\
         \bottomrule
    \end{tabular}}
    \caption{KilometerAudio results. $^\dagger$Best-performing entry, not eligible for a prize; see text.}
    \label{tab:kawin}
\end{table}

\subsection{Best Generalist award}
The Best Generalist award was received by Wind\_Rain\_Tower team, who submitted to all four tracks with a single training-free approach. The Qwen3.8-Max model makes every final answer and box decision, without task-specific detectors, trackers, or geometry solvers. Only a separate speech-recognition model is used to transcribe the audio for KilometerAudio as the model cannot process audio natively. The results of the generalist model were: 0.896 top-1 in multiple-choice video QA (5th place, 98\% relative to the top solution), 0.465 HOTA in grounded video QA (13th place, 69\% of top), 0.604 in KilometerVision (9th place, 80\% of top), and 0.506 in KilometerAudio (22nd place, 65\% of top). Although the model trailed behind the per-track winners, this is the first submission across the four editions that shows a unified model with satisfactory performance across all the tracks, giving hope that a complete perception model is within reach.   

\section{Discussion}

The Fourth Perception Test challenge attracted a record number of submissions (1254 test submissions) from 120 teams across four tracks. Except for the Grounded VideoQA track, the performance on the remaining three tracks almost saturated ($>80\%$ accuracy). This is remarkable, especially given that two of the tracks, the city-scale tracks, have been run this year for the first time. 


Almost all top solutions were zero-shot agentic pipelines built around proprietary or open-source VLMs (Gemini, GPT, Doubao-Seed, Qwen), often combined with ensembling or self-consistency metrics. Among the awarded entries, only the unified multiple-choice winner fine-tuned a model on the Perception Test training data. The best performing methods in the new city-scale tracks relied on expensive and slow agentic tool use, audio indexing, or explicit geometric reconstruction, but obtained almost perfect performance. In contrast, the Best Generalist entry, which used a single multimodal model for all tasks, stayed close to the winners on multiple-choice video QA, but trailed behind on the city-scale tracks. In conclusion, large-scale audio-visual and spatial reasoning can be solved by systems that orchestrate models with explicit tools, but remains an open problem for multimodal models used standalone.

\subsection*{Acknowledgements} We thank Google DeepMind for funding the prizes and swag, and the EvalAI team for their help in hosting the four tracks.

\bibliography{main}

\end{document}